\newcommand{\CLASSINPUTtoptextmargin}{19mm}
\newcommand{\CLASSINPUTbottomtextmargin}{18mm}
\newcommand{\CLASSINPUTinnersidemargin}{19mm}
\newcommand{\CLASSINPUToutersidemargin}{19mm}

\documentclass[letterpaper, conference]{IEEEtran}

\usepackage[utf8]{inputenc}
\usepackage{xcolor}
\usepackage[english]{babel}
\usepackage{amsmath}
\usepackage{csquotes}
\usepackage[parfill]{parskip}
\usepackage{color}
\usepackage{placeins}
\usepackage{hyperref}
\usepackage{subcaption}
\usepackage{flushend}

\usepackage{graphicx}

\title{\vspace{6mm}Towards Adaptive Benthic Habitat Mapping}
\author{\IEEEauthorblockN{Jackson Shields, Oscar Pizarro, Stefan B. Williams}
\IEEEauthorblockA{Australian Centre for Field Robotics, University of Sydney, Sydney, NSW Australia}
\{j.shields, o.pizarro, stefanw\}@acfr.usyd.edu.au
}
\date{August 2019}

\usepackage[maxcitenames=2,uniquelist=false]{biblatex}

\begin{document}
\maketitle
\begin{abstract}

Autonomous Underwater Vehicles (AUVs) are increasingly being used to support scientific research and monitoring studies.  One such application is in benthic habitat mapping where these vehicles collect seafloor imagery that complements broadscale bathymetric data collected using sonar. Using these two data sources, the relationship between remotely-sensed acoustic data and the sampled imagery can be learned, creating a habitat model. As the areas to be mapped are often very large and AUV systems collecting seafloor imagery can only sample from a small portion of the survey area, the information gathered should be maximised for each deployment. This paper illustrates how the habitat models themselves can be used to plan more efficient AUV surveys by identifying where to collect further samples in order to most improve the habitat model. A Bayesian neural network is used to predict visually-derived habitat classes when given broad-scale bathymetric data. This network can also estimate the uncertainty associated with a prediction, which can be deconstructed into its aleatoric (data) and epistemic (model) components. We demonstrate how these structured uncertainty estimates can be utilised to improve the model with fewer samples. Such adaptive approaches to benthic surveys have the potential to reduce costs by prioritizing further sampling efforts.  We illustrate the effectiveness of the proposed approach using data collected by an AUV on offshore reefs in Tasmania, Australia.

\end{abstract}

\section{Introduction}

Marine scientific surveys are conducted in support of a range of marine science research including ecology, geology and archaeology~\cite{Williams2012MonitoringVehicle},\cite{Snelson2005Sparse},\cite{Williams2016ReturnSite}. Autonomous Underwater Vehicles (AUVs) are crucial for conducting these surveys, as they can collect data in areas that are inaccessible to humans, they have greater endurance and they can collect a wealth of data using a range of onboard sensors~\cite{Singh2004SeabedImaging}. Recent years have seen AUV systems being increasingly used to collect seafloor imagery that complements broadscale bathymetric data collected by ships or high-flying AUVs.  However, due to the strong attenuation of electromagnetic radiation in water, visual data has to be collected close to the target, resulting in a relatively small sensor footprint. Even given the enhanced endurance of current generation AUVs, the area they can visually sample from is limited by this small sensor footprint, driving the need to conduct efficient surveys that maximise the information collected within a given target area of the seafloor.

\begin{figure*}[!htbp]
	\centering
    \includegraphics[width=0.8\textwidth]{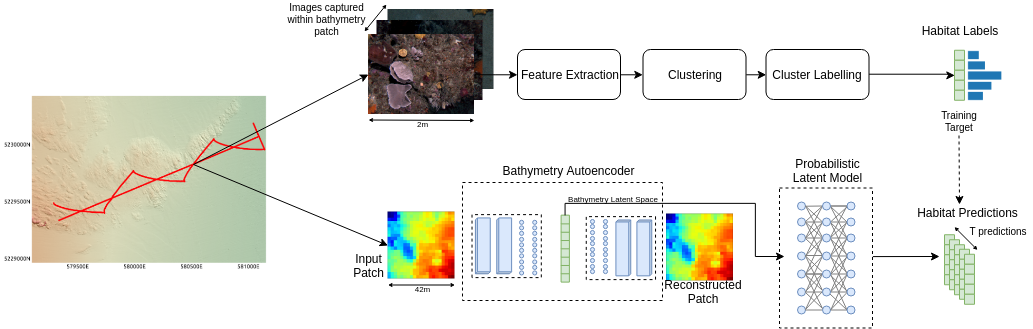}
    \caption{Process Diagram.  (Top row) Images are first clustered and then consolidated by an expert to yield a small set of ecologically relevant class labels. As the bathymetric patches are significantly larger than the footprint of an image, the distribution of habitat labels within each patch is calculated. (Bottom row) Bathymetric patches are fed through an autoencoder that yields a latent space used to train a Bayesian neural network.  The network is trained to classify unseen patches using the training targets provided by the image-based habitat labels.}
    \label{process:diagram}
\end{figure*}

At present, AUV surveys are mostly planned manually, with the survey planner inspecting remotely-collected data, such as bathymetry and/or backscatter to identify areas of interest and designing a path that visits these regions.  However, these plans do not explicitly take into account the mapping process itself in order to identify the most efficient locations to sample.  This paper focuses on building probabilistic habitat models with minimal human supervision that can be utilised to plan efficient surveys. These models map the remotely-sensed data (bathymetry) to the habitat class, which is estimated from imagery collected by the AUV. The system developed (see Figure \ref{process:diagram}) can be utilised to plan efficient surveys and therefore maximise the utility of AUV deployments. 

The key contributions of this paper are: 
\begin{enumerate}
    \item The development of predictive habitat models with minimal human supervision.
    \item The application of Bayesian neural networks to habitat modelling, providing probabilistic predictions and scalability.
    \item Analysis of deconstructed uncertainties to identify areas which should undergo further sampling.
    \item Demonstration of using epistemic uncertainty for active learning.
\end{enumerate}

The remainder of this paper is organised as follows: Section \ref{Sec:RelatedWork} provides an overview of related literature, focusing on remote-sensing of benthic habitats; Section \ref{Sec:Method} details the pipeline used for remote habitat modelling; Section \ref{Sec:Data} provides an overview of the data used in this paper; Section \ref{Sec:Results} presents the results from this method, while Section \ref{Sec:Discussion} explores active learning with the habitat model. Finally, Section \ref{Sec:Conclusion} provides a conclusion and identifies avenues of future research.

\section{Related Work}\label{Sec:RelatedWork}
For marine habitat-modelling, the remotely-sensed data used is typically bathymetry and backscatter, collected from ship-borne sonars~\cite{Brown2011BenthicTechniques}. More recently, hyperspectral data from satellites and Lidar deployed from small aircraft has also been utilised for habitat modelling in shallow waters~\cite{Ohlendorf2011BathymetryProcessing}.

\textcite{Dartnell2004PredictingData} collected sediment samples and used decision trees to relate these samples to the acoustic data. However, collecting sediment samples is time intensive and instead the benthic habitat type can be estimated from seafloor imagery~\cite{Brown2011BenthicTechniques}. \textcite{Kostylev2001BenthicPhotographs} collect benthic imagery using a drop camera, which allows data collection in areas inaccessible to humans. The habitat types inferred from this imagery are related to the terrain complexity, depth, water current and backscatter to produce a habitat map. This method is limited by the slow rate of data acquisition when using drop cameras, resulting in sparse samples. Benthic imaging AUVs equipped with advanced navigation solutions collect georeferenced imagery as they run their survey paths~\cite{Singh2004SeabedImaging}, thereby greatly increasing the samples collected. The expanded data collection using these vehicles allows the use of data-driven machine learning models.

\textcite{Marsh2009NeuralIV} use self-organising maps to train an unsupervised bathymetry and backscatter classifier. This classifier groups similar areas of bathymetry and backscatter, however it has no notion of the habitat classes present in these groups, and further classification is needed to predict the habitat class. \textcite{Ahsan2012} use Gaussian Mixture Models (GMMs) to predict the habitat class from bathymetry, by first extracting the morphological features of rugosity, slope and aspect. An emphasis is placed on using predicted entropy to direct survey planning. \textcite{Bender2012ClassificationTargets} utilise Gaussian Processes (GPs) to model the habitat classes from bathymetric features. The habitat classes are estimated from clusters derived from the AUV imagery~\cite{Steinberg2012AnData}. The benefit of using GPs is the probabilistic output, allowing the estimation of uncertainty, which can be used to direct future surveys to reduce uncertainty in the model. However, GPs do not readily scale to increased dataset sizes, which is necessary for a dataset to generalise over a larger area with different characteristics and to take advantage of large image datasets. To provide both scalability and a probalistic output, this work utilises Bayesian Neural Networks (\cite{Neal1996}). \textcite{Rao2017} utilise a multimodal mixture Restricted Boltzmann Machine (RBM) to model the joint distribution between the benthic imagery and the bathymetry. This model allows classification based on either or both modalities. Rather than extracting morphological features from the bathymetry (rugosity, slope, aspect), a denoising autoencoder is used for feature extraction. This allows more information in the feature space and is adaptable to other remote data types, promoting its use in this research.

\section{Method}\label{Sec:Method}
Figure \ref{process:diagram} illustrates the method proposed in this study.  Habitat labels derived from imagery are used to train a Bayesian neural network using the latent space of an autoencoder that can reconstruct the bathymetric patches.  The steps in this process are detailed below.  

\subsection{Habitat Classes From Image Clusters}

The habitat labels are defined based on the imagery~\cite{Brown2011BenthicTechniques}. An AUV can capture tens of thousands of seafloor images per dive (deployment), which results in hundreds of thousands of seafloor images of the survey area. Manually labelling these images is infeasible, and as such the feature extraction and clustering pipeline developed by \textcite{Steinberg2012AnData} is used to assign habitat labels with minimal human supervision. Image features are extracted using the ScSPM (Sparse code Spatial Pyramid Matching)~\cite{Yang2009LinearClassification}, which are then clustered using the GMC clustering method ~\cite{Steinberg2011ASurveys}. After clustering is completed, the clusters are further grouped together to form habitat classes, based on examples from each cluster.

\subsection{Bathymetric Feature Extraction}
Bathymetry is strongly correlated to habitat class~\cite{Friedman2012}, with the features of depth, slope, aspect and rugosity being key indicators for this relationship. Extracting these features at different scales can lead to improved performance for habitat classification~\cite{Bender2013}. A limitation of extracting these hand-picked features is that they represent complex data sources in too few features, even if extracted at different scales. This limits the amount of information the habitat model is able to use. Alternatively, learnt features can be used. Convolutional neural networks (CNNs) have provided breakthroughs in image processing, as they are capable of extracting rich features from the images~\cite{Lecun2015}. These same techniques are applicable to the bathymetry. To create a useful feature extractor from the CNN without labelled data, an autoencoder is used. An autoencoder consists of two modules; an encoder and a decoder. The encoder processes the input and outputs a compressed representation, while the decoder takes in this compressed representation and aims to recreate the input. After training, the compressed representation is a set of features that efficiently explain the data.  The autoencoder is able to reconstruct bathymetry patches with little error ($0.05m$ mean squared error), meaning the entire bathymetry patch is compressed into the latent dimensions. This approach has been used to provide features for simple classifiers, which achieve high accuracy with limited data~\cite{Le2013BuildingLearning}, highlighting the quality of the learnt features. Furthermore, this approach is versatile and can be applied to any remote data including backscatter or hyperspectral data.

The encoder utilises two convolutional layers each with 1024 and 512 filters respectively, and a kernel size of 3. Following this, two fully connected layers are used with 512 units in each. The latent space was 32 dimensions when aiming to reconstruct a 21x21 raster patch. The decoder has a mirrored network structure, with an additional single filter convolutional layer to output the reconstructed patch. 

\subsection{Latent Mapping}
Latent mapping estimates the habitat class given the latent space from the bathymetry and/or backscatter. As displayed in Figure \ref{process:diagram}, the spatial extents of the bathymetry differ by one or two orders of magnitude. Each bathymetry patch is 42m by 42m, whereas an image typically has a footprint of 1.5m by 1.2m. There is a many-to-many relationship between bathymetry and images; a bathymetry patch can contain many different habitat types, whereas a habitat type can belong to many different types of bathymetry patches. A probabilistic model is a natural fit for this relationship. When sampling from the model, the distribution of the samples should reflect the underlying distribution of habitats given that bathymetry patch. The uncertainty associated with a prediction can also be estimated, which is vital for the model to be part of an active learning pipeline, where the aim of sampling is to improve the model.


As a Bayesian neural network uses Monte Carlo sampling to create a probabilistic output, the model needs to be sampled multiple times, which motivates the need for efficient models. Using a model that inputs the latent space rather than raw data is more efficient, as feature extraction does not need to be performed for each sample.

\subsection{Bayesian Neural Networks}
\textcite{Bender2012ClassificationTargets} used the bathymetric features of rugosity, slope, aspect and depth as input to a GP classifier which was used to predict the habitat class. A limitation of using GPs is that they scale poorly with the number of training data points~\cite{Hensman2015ScalableClassification}. Computationally attractive special cases and approximates are still an active area of research (\cite{Cutajar2017RandomProcesses},~\cite{Krauth2017AutoGP:Models}). The latent model needs to be able to learn from all the collected imagery (millions of data points). Bayesian Neural Networks (BNNs) are alternatives to GPs which can be used to make predictions with uncertainty. A BNN, as popularised by \textcite{Neal1996}, is a feedforward neural network, where a distribution is placed over each of the parameters (weights and biases) of the network. BNNs have two significant advantages over GPs; they explicitly handle multi-class / multi-output targets and they scale to large datasets.

Exact inference over complex Bayesian graphs (such as a BNN) generally involves intractable integrals in the posterior. To approximate the posterior, approximation methods are used. MCMC (Markov Chain Monte Carlo~\cite{Hastings1970MonteApplications}) uses sampling to estimate the posterior, however, for complex models sampling is computationally intensive. For large datasets and complex models, Variational Inference (VI)~\cite{Jordan1999AnModels} is a viable alternative for approximate inference. 

The underlying idea behind Variational Inference is that a simplified approximate distribution $q$ with variational parameters $\theta$ is used to approximate the actual distribution $p$~\cite{Blei2017VariationalStatisticians}. \textit{Bayes by Backprop}~\cite{Blundell2015WeightNetworks} utilises variational inference to adapt backpropagation to train a BNN. For a BNN where the approximate distribution $q$ is placed over the weights of the network $w$ and subject to training data $D$~\cite{Blundell2015WeightNetworks}:
\begin{equation}
q_{\theta} (w | D) \approx p (w | D) 
\end{equation}

The approximate distribution is calculated by minimising the KL divergence between $p$ and $q$, which is framed as an optimisation problem. This is commonly known as the expectation lower bound (ELBO)~\cite{Blundell2015WeightNetworks}:
\begin{equation}
\theta^{*} = arg _\theta min KL [q_\theta (w | D ) || P (w | D)]   
\end{equation}
However the KL divergence is also intractable, as it involves a complex integral. \textit{Bayes by Backprop} uses sampling to approximate the KL divergence. First it approximates the distribution $q$ by learning its parameters and then samples from this approximate distribution $q$ given the data. This is summarised in the optimisation objective~\cite{Blundell2015WeightNetworks}:
\begin{equation}
    \theta^{*} = \sum^n_{i=1} \text{log}\ q(w^{(i)} | \theta) - log\ P(w^{(i)}) - log\ P (D | w^{(i)})
\end{equation}
where $w^{(i)}$ represents a Monte Carlo sample drawn from the approximate posterior $q(w^{(i)}|\theta)$. To provide gradient estimates for these distributions, \textit{Bayes by Backprop} utilises the reparameterization trick from~\cite{Kingma2015VariationalTrick}. 

In this paper, the BNN was implemented using PyTorch~\cite{Pazke2017AutomaticProse} and Pyro~\cite{Bingham2019Pyro:Programming}. It consists of 3 fully connected layers, all with 2048 units, followed by the logits layer. The approximate distribution is parameterized by Gaussian distributions.

\subsection{Deconstructing Uncertainty}


The uncertainty estimate of a Bayesian neural network can stem from either uncertainty in the data (aleatoric) or uncertainty in the model (epistemic). Using the BNN approximation technique Monte Carlo dropout~\cite{Gal2016DropoutLearning}, \textcite{Kendall2017} deconstruct the uncertainty into its epistemic and aleatoric components. A multi-head network is used, where one head is the predicted mean (class vector) and the other is the predicted variance. Both the predictive mean and predictive variance are trained, where the predictive variance is learned implicitly, not requiring uncertainty labels. \textcite{Kwon2018} build upon~\cite{Kendall2017}, by deconstructing the uncertainty without using an extra network head.

The total uncertainty is given by~\cite{Kwon2018}:
\begin{equation}
        \text{Var} = \underbrace{ \frac{1}{T} \sum^{T}_{t=1}\text{diag}(\hat{y}_t) - \hat{y}_t^{\otimes 2} }_\text{aleatoric} + \underbrace{ \frac{1}{T} \sum^T_{t=1} (\hat{y}_t - \bar{y})^{\otimes 2}}_\text{epistemic} 
\end{equation}
Where $\hat{y}_t$ is the prediction for each sample, $t$, of the model and $^{\otimes}$ is the tensor-wise product. The deconstructed uncertainties provide insight into how to sample next. As epistemic uncertainty highlights areas of model uncertainty, further sampling of these areas can improve the model, enabling a more comprehensive habitat model to be formed with less sampling. Aleatoric uncertainty represents noise in the data and cannot be reduced by further sampling.

\section{Data}\label{Sec:Data}

\begin{figure}[!htbp]
	\centering
    \includegraphics[width=\columnwidth]{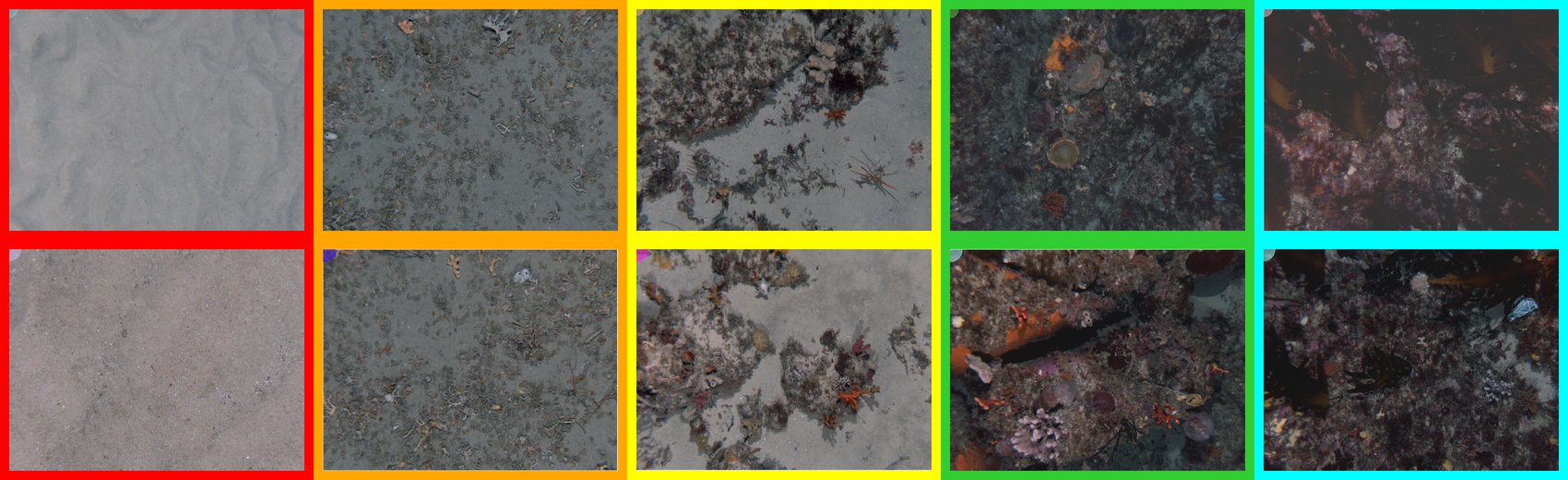}
    \caption{Habitat classes found in the Tasmania data. From left to right: sand, screwshell rubble, patchy reef, reef, kelp. The border colour indicates the plotting colour (Figure \ref{results:bayesnn:categories}).}
    \label{data:tasmania:classes}
\end{figure}

The area of focus is the Fortescue region of Tasmania, with the dataset containing ~149535 geo-referenced images collected in October 2008, as well as bathymetry collected by GeoScience Australia~\cite{Spinoccia2011BathymetryShelf} with a ship-borne multibeam.
The habitat categories found in this dataset are Sand, Screwshell Rubble, Patchy Reef, Reef and Kelp. Example images are displayed in Figure \ref{data:tasmania:classes}. There is a significant difference between the resolution of the imagery and the bathymetry. The bathymetry is gridded at 2m, meaning that each patch is 42m by 42m (as each patch is 21x21 cells). This contrasts to the imagery, which typically has a footprint of $~$2m. This difference in resolution motivates the decision to represent the habitat map as a distribution of class labels per bathymetry patch.


\begin{figure}[!htbp]
    \centering
    \begin{subfigure}[t]{0.8\columnwidth}
        \centering
        \includegraphics[width=\textwidth]{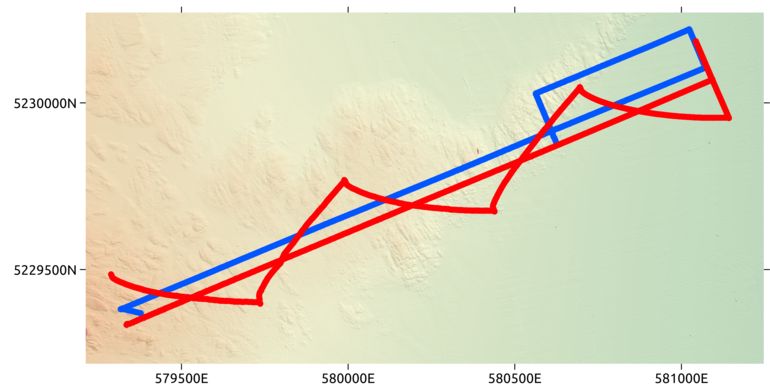}
        \vspace*{-5mm}
        \caption{Ohara}
        \label{data:tasmania:ohara}
    \end{subfigure}
    ~ 
    \begin{subfigure}[t]{0.8\columnwidth}
        \centering
        \includegraphics[width=\textwidth]{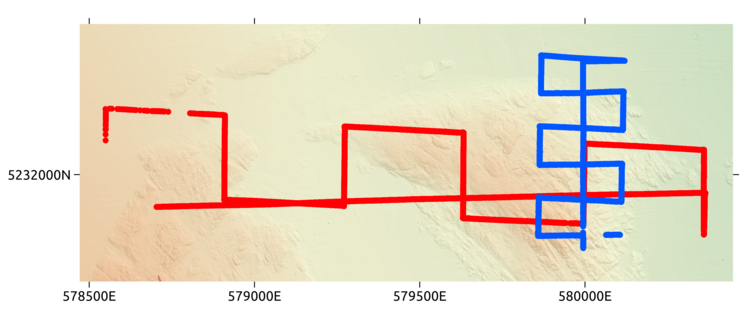}
        \vspace*{-5mm}
        \caption{Waterfall}
        \label{data:tasmania:waterfall}
    \end{subfigure}
    \caption{Bathymetry of two study sites within the Fortescue region of Tasmania. Imagery location is indicated by the AUV track overlays with red for training and blue for validation. For (\ref{data:tasmania:ohara}), Ohara 07 is training and Ohara 20 is validation. For (\ref{data:tasmania:waterfall}), Waterfall 05 is training and Waterfall 06 is validation. UTM (Universal Transverse Mercator) Zone 55S projections. }
    \label{data:tasmania:map}
    \vspace*{-3mm}
\end{figure}

\subsection{Data Splitting}

Since habitat extents are typically much larger than the footprint of an image, care should be taken for dataset splitting. A na{\"i}ve training/validation split randomly assigns data points to each set. However, this is problematic for spatial information where neighbouring points can be very similar. For example, the bathymetry patches centered around two sequential images collected by the AUV are nearly identical, just shifted by as little as one metre. Since neighbouring data points often have the same habitat class, if these data points are randomly assigned to training and validation sets, validation will not be indicative of its generalisation.

An alternative approach is to split via dive (deployment), where one set of dives are used for training and another set for validation. To avoid data contamination, data points with overlapping bathymetry patches are removed from the validation set.

\section{Results}\label{Sec:Results}
\subsection{Metrics}

The Tasmania data was trained on Ohara 07 and Waterfall05 dives, and validated on Ohara 20 and Waterfall 06 (See Figure \ref{data:tasmania:map}). The datasets were balanced and all validation data points overlapping the training dives were removed. Three accuracy metrics are reported. The label accuracy, which is the rate of correct label predictions given the bathymetry patch. The neighbour accuracy, which is the rate of correct predictions of the modal class of the bathymetry patch. The benchmark accuracy, which is the rate that the habitat label aligns with the modal class of the bathymetry patch. This measurement is regarded as the benchmark accuracy as it reflects what a near perfect habitat classifier would achieve, as the best it can be expected to achieve is to predict the modal class. The model reported a label accuracy of 0.60, neighbour accuracy of 0.70, while the benchmark accuracy was 0.74.

\begin{figure}[t]  
    \centering
    \includegraphics[width=0.57\columnwidth]{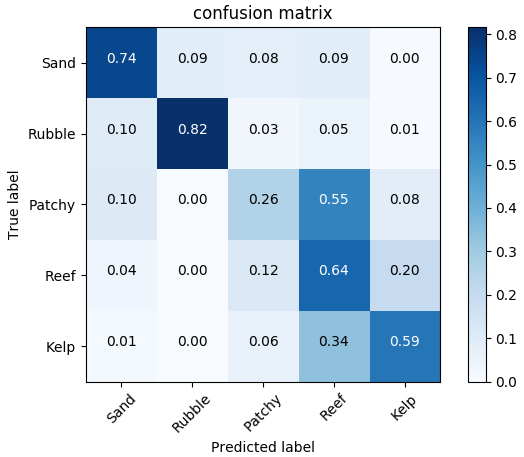}
    \caption{Confusion matrix for habitat label prediction}
    \label{results:bayesnn:metrics:confusion}
\end{figure}

Trying to predict a fine-scale habitat class based on low resolution data is difficult, as highlighted by the difference between the label accuracy and the neighbour accuracy. Although it was not directly trained to predict the modal class, the classifier performs well on the neighbour accuracy task. As demonstrated in Figure \ref{results:bayesnn:metrics:confusion} the patchy label is confused primarily with reef, which is understandable since the transition occurs at scales smaller than a single bathymetry patch. There is also confusion between kelp and reef, which is again understandable given that kelp grows on reef and mostly does not appear in the acoustic data.



\begin{figure}[t] 
    \centering
    \includegraphics[width=0.57\columnwidth]{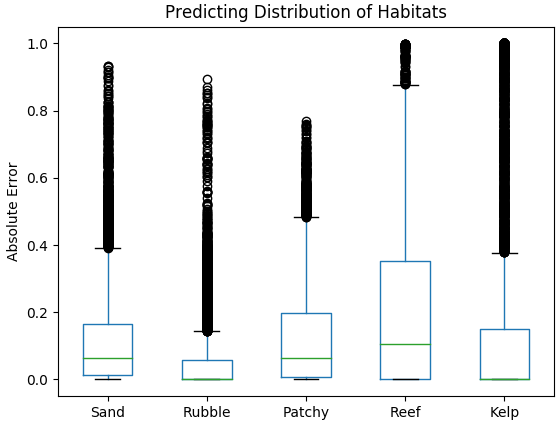}
    \caption{Absolute error box plot for predicting the overall distribution of samples within a bathymetry patch. The green line depicts the median, while the edges of the box depict the $1^{st}$ and $3^{rd}$ quartiles. The whiskers are defined as $1.5 IQR$ (Interquartile Range) from the edges of the box.}
    \label{results:bayesnn:metrics:distribution_error}
\end{figure}

Providing a prediction of the distribution of habitats within a given bathymetry patch is more appropriate than a single point estimate, given the large scale difference between bathymetry and images. When sampling from the model, the distribution of these samples should match the underlying distribution of habitat classes within the patch. Figure \ref{results:bayesnn:metrics:distribution_error} shows the mean absolute error between the sampled distribution and the underlying distribution. The low error demonstrates that sampling from the model reflects the underlying distribution of habitat classes.


\subsection{Maps}

\begin{figure}
    \centering
    \begin{subfigure}[t]{0.8\columnwidth}
        \centering
        \includegraphics[width=\textwidth]{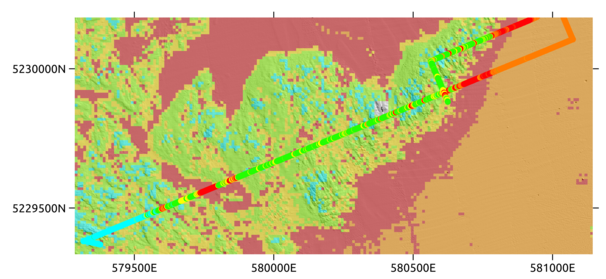}
        \caption{Ohara}
        \label{results:bayesnn:categories:ohara}
    \end{subfigure}
    ~ 
    \begin{subfigure}[t]{0.8\columnwidth}
        \centering
        \includegraphics[width=\textwidth]{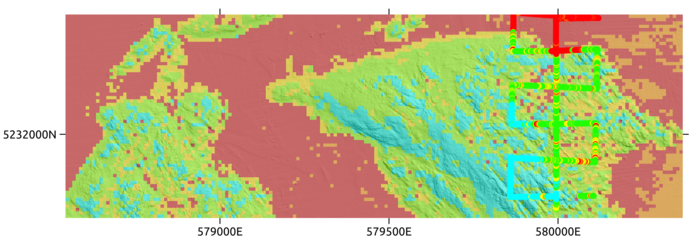}
        \caption{Waterfall}
        \label{results:bayesnn:categories:waterfall}
    \end{subfigure}
    \vspace*{-1mm}
    \caption{Predicted categories for the Ohara and Waterfall regions. The AUV track overlay is the validation dive for each region. Colour labels: red=sand, orange=rubble, yellow=patchy, green=reef, cyan=kelp. UTM Zone 55S projection.}
    \label{results:bayesnn:categories}
\end{figure}

The predicted category maps are displayed in Figure \ref{results:bayesnn:categories}. The validation track overlay highlights a high degree of correlation between the predicted habitat class and the image labels. The predictions also correlate with domain knowledge of this area. The shallower ($<45m$) rugged areas are more likely to feature kelp habitats, as kelp requires light to grow. Deeper rugged areas are often characterised as reef. The interface between the reef/kelp areas and the sand areas are characterised as patchy reef. The deeper flatter sections are classified as screwshell rubble, which is a feature of these areas.

\begin{figure}[!htbp]
    \centering
    \begin{subfigure}[t]{0.8\columnwidth}
        \centering
        \includegraphics[width=\textwidth]{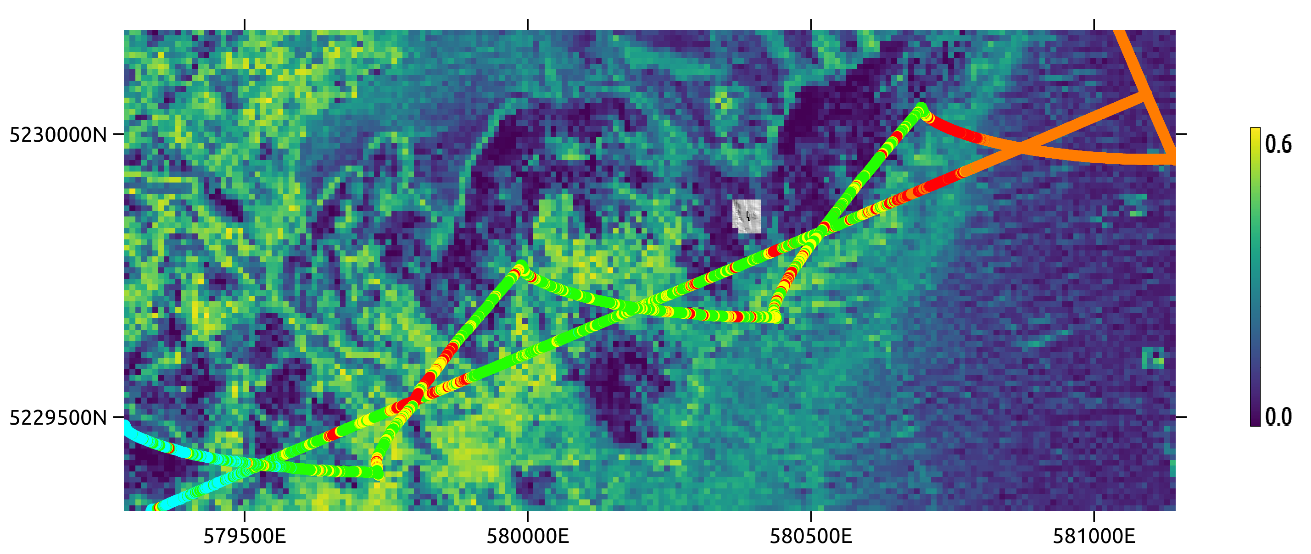}
        \caption{Aleatoric uncertainty: Ohara}
        \label{results:bayesnn:aleatoric:ohara}
    \end{subfigure}
    ~ 
    \begin{subfigure}[t]{0.8\columnwidth}
        \centering
        \includegraphics[width=\textwidth]{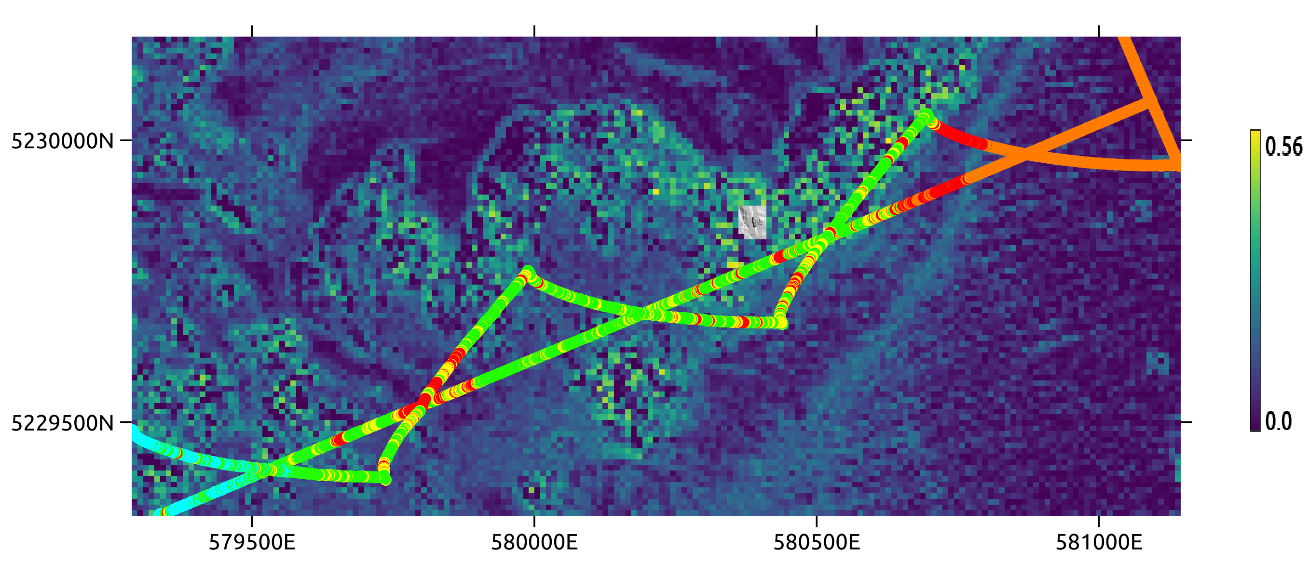}
        \caption{Epistemic uncertainty: Ohara}
        \label{results:bayesnn:epistemic:ohara}
    \end{subfigure}
    \vspace*{-1mm}
    \caption{Aleatoric and epistemic uncertainties for the Ohara region. The track overlays are the training dive, to show the areas it was trained on. UTM Zone 55S projection.}
\end{figure}

\subsection{Uncertainty}
The aleatoric uncertainty map (Figure \ref{results:bayesnn:aleatoric:ohara}) highlight areas of high data uncertainty, where similar bathymetry patches can have a wide distribution of labels. It is higher around interface areas between reef and sand, where it is difficult to differentiate between classes using the low-resolution bathymetry.

The epistemic uncertainty maps (Figure \ref{results:bayesnn:epistemic:ohara}) display the areas of model uncertainty. Emphasis is placed upon the reef outcrops, particularly those with a Northerly aspect, which are under-represented in the training data. The model identifies it is confident in predicting flat areas, as there is low uncertainty in the flat areas to the East of the Ohara area. 


\section{Active Learning with Uncertainty}\label{Sec:Discussion}

Using models that provide epistemic uncertainty estimates provide an avenue towards active learning, where the objective is to improve the model with fewer additional training points. However there is no ground truth for uncertainty, making it hard to validate. To evaluate the use of epistemic uncertainty in active learning, a selective training experiment is proposed, which aims to mimic an adaptive sampling procedure. A model is trained on a small portion of the data, and at each iteration selects what samples to train on next, based on either the epistemic uncertainty, aleatoric uncertainty or selected at random. The hypothesis is that using the epistemic uncertainty to select where to sample will enable the model to reach the performance limit with fewer training points.


\begin{figure}
    \centering
    \begin{subfigure}[t]{0.2\columnwidth}
        \centering
        \includegraphics[width=\textwidth]{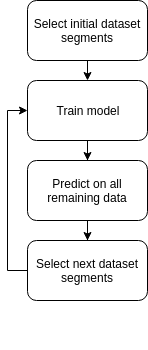}
        \caption{}
        \label{discussion:selective:diagram}
    \end{subfigure}
    ~ 
    \begin{subfigure}[t]{0.74\columnwidth}
        \centering
        \includegraphics[width=\textwidth]{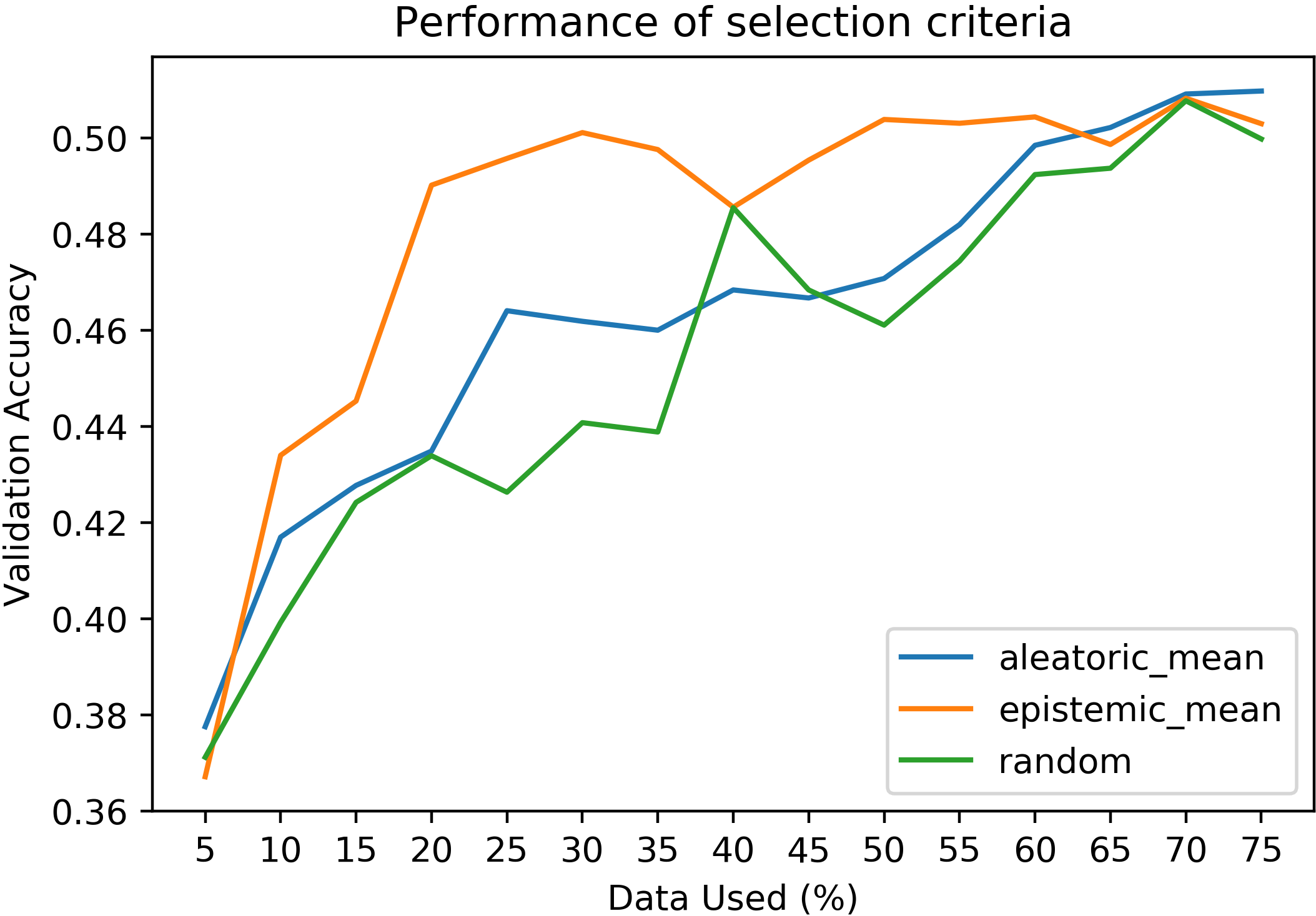}
        \caption{}
        \label{discussion:selective:valplot}
    \end{subfigure}
    \vspace*{-2mm}
    \caption{(a) The selective training process. (b)~The mean validation label accuracy at each iteration for each selection criteria, when performing selective training as validated on the Ohara 20 dive (see Figure \ref{data:tasmania:ohara}). Results are averaged over three runs and illustrate how the epistemic sampling shows faster convergence of the validation accuracy suggesting that it is picking areas to sample that are more effective at reducing model uncertainty. }
\end{figure}

The selective training process is summarised in Figure \ref{discussion:selective:diagram}. First, the Ohara 07 training dataset (see Figure \ref{data:tasmania:ohara}) is split into 100 sequential segments. Each one of these segments represents an area to be potentially sampled. The model is trained for 10 epochs, which is enough for it to converge. Next, inference is run on all the remaining dataset segments, and the epistemic and aleatoric uncertainties are calculated. Based on the prediction results, a further 5 dataset sections are selected for training, based on one of the following criteria: the maximum mean epistemic uncertainty; the maximum mean aleatoric uncertainty, or selected at random. This process of training, predicting, then selecting is repeated for 15 iterations. At 15 iterations, 80\% of the data is selected, at which point the model performance usually converges to its maximum accuracy. 

Figure \ref{discussion:selective:valplot} displays the average validation accuracy for the three selection criteria. Epistemic uncertainty consistently outperforms the aleatoric uncertainty or random criteria, highlighting the applicability of epistemic-driven sampling. There are several factors which reduce the effectiveness of epistemic-based sampling in this experiment. There is significant label noise in many areas and training on these segments can introduce further confusion. The similarity between classes can make the model certain about its prediction if it has not seen the other class. For example, the patches containing sand and rubble are both relatively flat and are very similar in the latent space. If the model has only been trained on one of these classes, it will be certain of its predictions of both the areas. Despite these drawbacks, epistemic uncertainty based selection offers the clearest path to model improvement. Using epistemic uncertainty to plan further sampling can produce more complete habitat models with fewer samples.

\section{Conclusion and Future Work}\label{Sec:Conclusion}



This paper demonstrates the application of Bayesian neural networks for building probabilistic habitat models. The focus has been on building these models with minimal human supervision. Convolutional autoencoders are used for feature extraction, as they can be applied to various remotely-sensed data sources and extract informative features. This research shows that using epistemic uncertainty to direct further samples results in greater model improvement with fewer samples, therefore making AUV surveys more efficient.

Future work will integrate this into online field trials with an AUV. This will involve running multiple deployments over a survey area and analysing whether sampling based on epistemic uncertainty can lead to a more comprehensive habitat model with fewer samples than current manually planned surveys. Using uncertainty to direct sampling opens up new challenges, including understanding how sampling from one area will impact the uncertainty in another.




\section{Acknowledgements}\label{Sec:Acknowledgements}
This work is supported by the Australian Research Council and the Integrated Marine Observing System (IMOS) through the Department of Innovation, Industry, Science and Research (DIISR) funded National Collaborative Research Infrastructure Scheme.  We also thank the Captain and crew of the R/V Challenger for their support of AUV operations. 

\printbibliography

\end{document}